%% file: acl_latex.tex
\documentclass[11pt]{article}

\usepackage[final]{acl}

\usepackage{times}
\usepackage{latexsym}

\usepackage{booktabs}       
\usepackage{xspace}
\usepackage{multirow}
\usepackage{ifthen}
\usepackage{arydshln}
\usepackage{tablefootnote}
\usepackage{graphicx}
\usepackage{subcaption}
\usepackage{amsmath}
\usepackage{cleveref}
\crefname{figure}{Fig.}{Figs.}
\crefname{table}{Tab.}{Tabs.}
\crefname{section}{Sec.}{Secs.}
\usepackage[table]{xcolor}
\usepackage{amssymb}

\input{preamble}

\newcommand{\name}{Argus-Unified\xspace}
\newcommand{\ie}{\textit{i.e.}\xspace}
\newcommand{\eg}{\textit{e.g.}\xspace}

\newcommand{\fakeparagraph}[1]{%
  \noindent\textbf{#1}\quad
}

\usepackage[T1]{fontenc}

\usepackage[utf8]{inputenc}

\usepackage{microtype}

\usepackage{inconsolata}

\usepackage{graphicx}

\title{Argus-Unified: Towards A Compact and Economical Unified Model for Image Understanding and Generation}

\author{Weiming Zhuang, Jiabo Huang, Jingtao Li, Zhizhong Li, \\ \textbf{Chen Chen, Sina Sajadmanesh, Lingjuan Lyu} \\
Sony AI \\
\texttt{\{weiming.zhuang, raymond.huang, jingtao.li, zhizhong.li\}@sony.com}\\
\texttt{\{chena.chen, sina.sajadmanesh, lingjuan.lv\}@sony.com}\\
}

\begin{document}
\maketitle

\begin{abstract}
Unifying visual understanding and generation in one model holds immense promise, but remains challenging and expensive due to heavy compute and data demands and conflicts between the visual features needed for these two capabilities.
To address these challenges, we present \ours,
a compact, effective and unified multimodal model built with low demand on computation and data.
Instead of aligning modalities from scratch,
\name effectively leverages pretrained vision-language models (VLMs) that provide strong multimodal priors.
Specifically, we introduce hybrid visual tokens that preserve continuous tokens for understanding while learning discrete tokens for generation from a frozen unified vision encoder.
Our training pipeline includes two stages: the first stage learns a quantizer and image decoder on top of the frozen vision encoder, the second stage trains the LLM initialized from a pretrained VLM for the unified multimodal modeling.
Using by far the least amount of data (15.6M) and the lowest cost ($\sim$\$2,000), we demonstrate that unified multimodal models can be trained economically while achieving strong performance in both understanding and generation. Notably, our model attains state-of-the-art multimodal understanding on GQA, POPE, and VQAv2, and competitive generation quality compared to models with dedicated vision encoders (e.g., Janus, Janus-Pro), all at \(\sim\)10\(\times\) lower cost and with \(\sim\)5\(\times\) less data.
We envision \name as a useful baseline that lowers the development barrier for unified models.
\end{abstract}

\begin{figure}[t!]
  \centering
  \begin{subfigure}[t]{0.45\textwidth}
      \includegraphics[width=\textwidth]{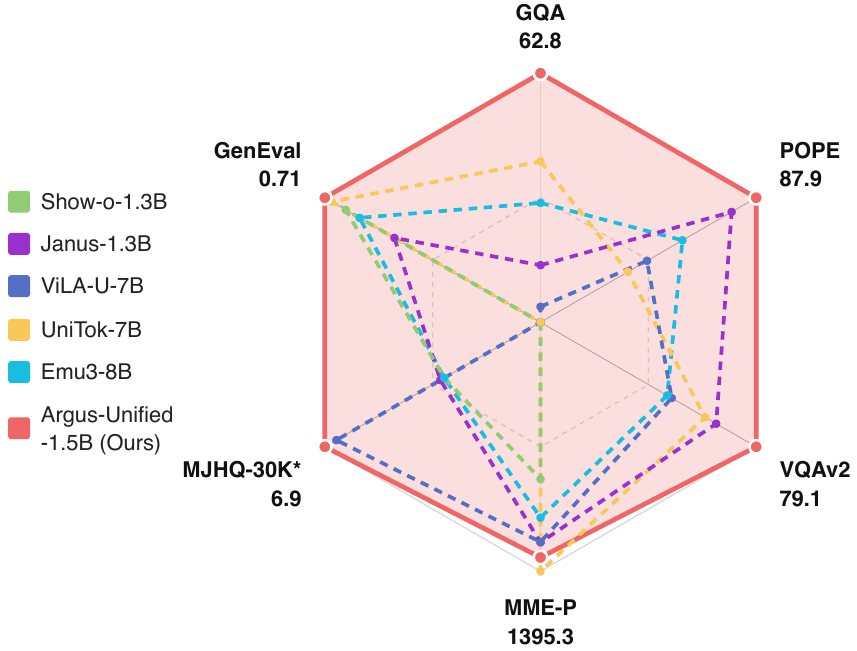}
  \end{subfigure}%
  \hfill
  \begin{subfigure}[t]{0.28\textwidth}
    \includegraphics[width=\textwidth]{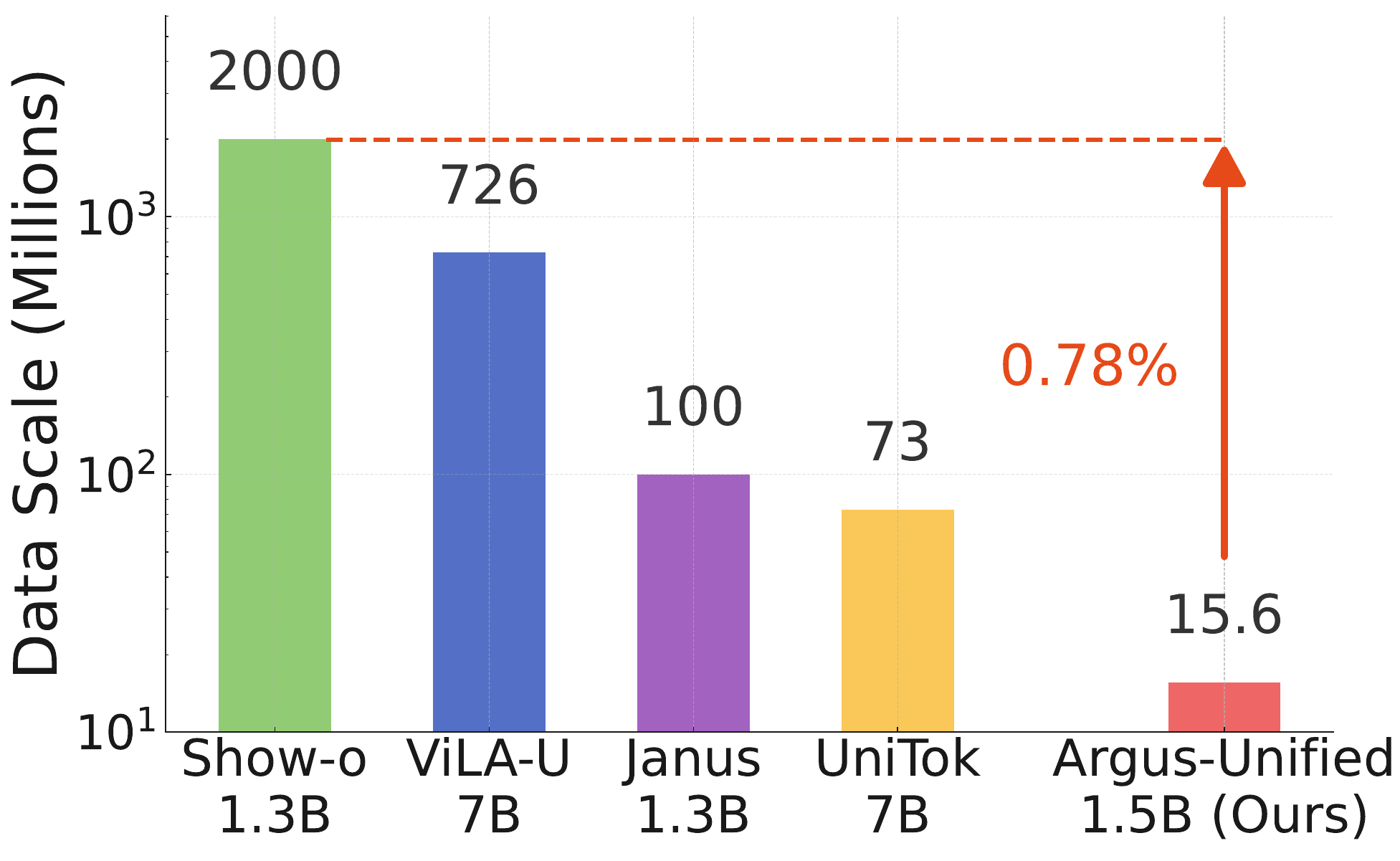}
  \end{subfigure}%
  \begin{subfigure}[t]{0.21\textwidth}
    \includegraphics[width=\textwidth]{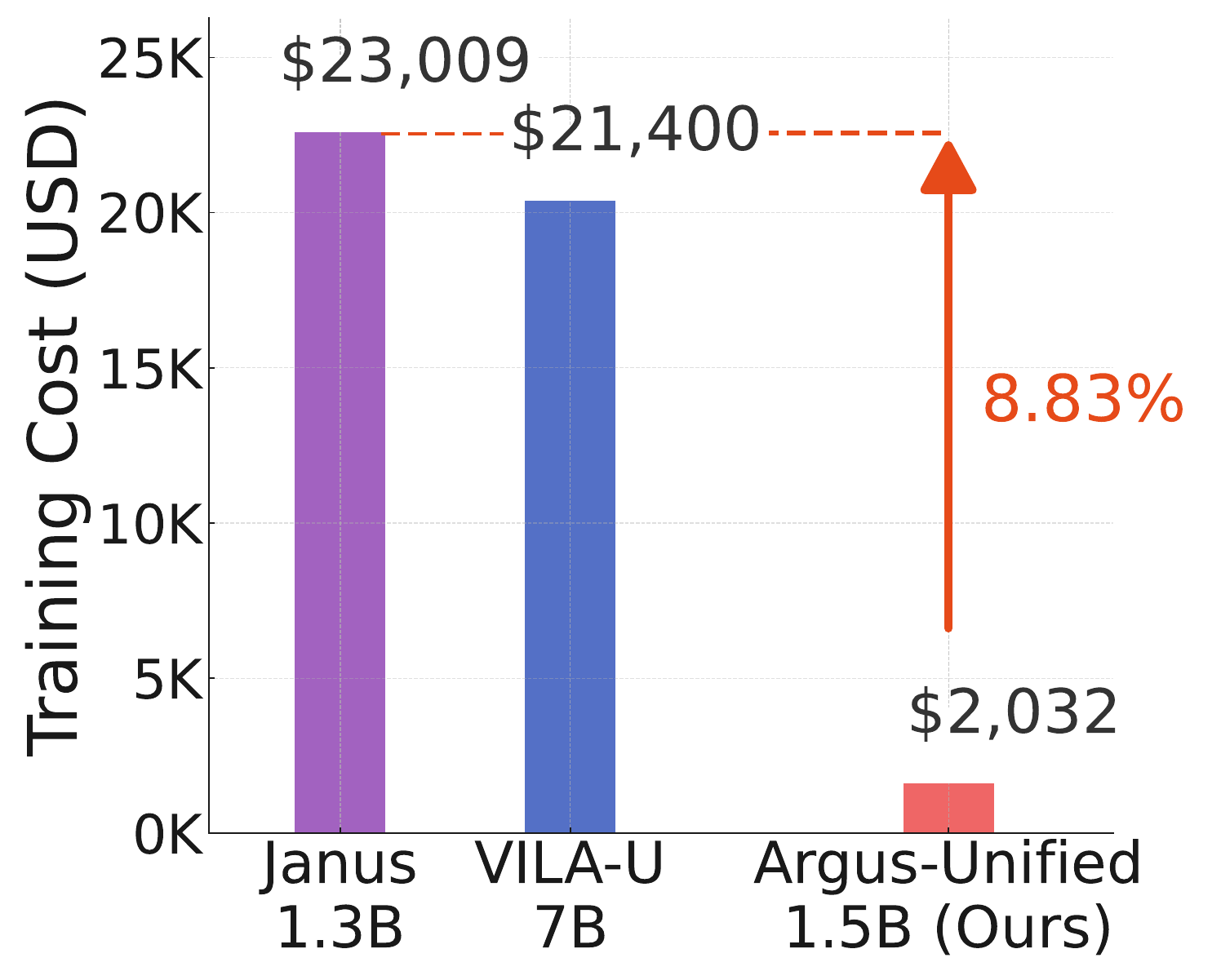}
  \end{subfigure}%
 \caption{\name{} achieves strong performance on both image understanding and generation across six benchmarks with only $\sim$\$2,000 economical cost and 15.6M data, substantially lower than existing unified multimodal models (UMMs). * denotes values inverted for illustration, where the original metric is lower-better.}
 \label{fig:performance}
\end{figure}

\section{Introduction}
\label{sec:intro}

With the success of large language models (LLMs) that unify natural language tasks~\cite{touvron2023llama,achiam2023gpt4,yang2025qwen3}, the pursuit at the forefront of unified multimodal models (UMM) that further integrates visual understanding and generation have become increasingly promising~\cite{team2024chameleon,wang2024emu3,wu2024vilau,wu2024liquid,ma2025unitok}.
Companies and institutions have built vision-language models (VLMs) for visual understanding based on pretrained LLMs~\cite{chen2024internvl2_5,lu2024deepseekvl,wu2024deepseekvl2,marafioti2025smolvlm,bai2025qwen2_5vl,zhu2025internvl3}.
Meanwhile, pioneering researches on visual generation has adopted autoregressive (AR) paradigm that predicts next visual tokens~\cite{li2024imagefolder,sun2024llamagen,tian2024var}.
These advances enable the unification of visual understanding and generation based on the AR paradigm.
However, existing UMMs primarily align modalities from scratch---relying on independently pretrained LLM and vision encoder
---without harnessing the multimodal priors in VLMs~\cite{wu2024vilau,wu2024liquid,ma2025unitok}.

\begin{figure*}[t!]
  \centering
  \includegraphics[width=\linewidth]{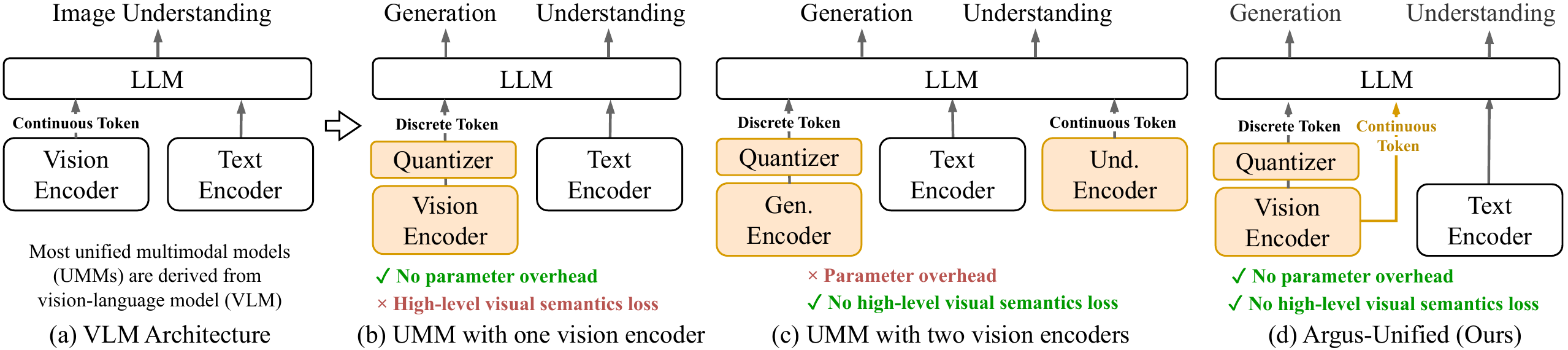}
  \caption{
Illustrations of model architectures of autoregressive vision-language model (VLM) and unified multimodal models (UMMs).
(a) Widely adopted VLM architecture that uses continuous tokens (i.e., features) for image understanding.
(b) Commonly adopted UMM architecture that uses a quantizer to produce discrete tokens for both understanding and generation based on a vision encoder~\cite{wu2024vilau, ma2025unitok,team2024chameleon,wang2024emu3,wu2024liquid}.
(c) Another popular UMM architecture that uses a generation (gen.) vision encoder with quantizer to produce discrete tokens and an understanding (und.) vision encoder for continuous tokens~\cite{wu2025janus,chen2025januspro,qu2025tokenflow,jiao2025unitoken}.
(d) We propose hybrid tokens, a new design for UMMs that leverages the best token types for respective tasks — continuous tokens for understanding and discrete tokens for generation, while keeping the architecture lightweight.
  }\label{fig:teaser}
\end{figure*}

Leveraging VLMs for unified multimodal models is not straightforward due to the tokenization difference between understanding and generation.
VLMs use \textit{continuous visual tokens} (\ie, visual features) from the vision encoder which contain rich semantic information (\cref{fig:teaser}(a))~\cite{liu2023llava,liu2024llava,zhu2025internvl3,bai2025qwen2_5vl}.
In contrast, autoregressive generation models commonly use \textit{discrete visual tokens} based on VQVAE~\cite{van2017vqvae} or VQGAN~\cite{esser2021taming}.
To address this mismatch, prior autoregressive UMMs mainly follow two approaches.
Some works~\cite{wang2024emu3,wu2024liquid,wu2024vilau,ma2025unitok} adopt discrete tokens for both understanding and generation (\cref{fig:teaser}(b)), at the cost of losing fine-grained semantic information critical for the understanding ability~\cite{wu2024vilau,ma2025unitok}.
The others ~\cite{wu2025janus,chen2025januspro,jiao2025unitoken} use two vision encoders, one for understanding and one for generation (\cref{fig:teaser}(c)).
They use continuous tokens for the understanding and discrete tokens from a separate vision tokenizer to fit the generation task.
However, this design increases model parameters with an additional encoder.

In light of this, we propose \name, a compact unified multimodal model that effectively leverages the pretrained VLMs to provide strong vision-language priors.
\name employs task-specific visual tokens from a single vision encoder:
continuous tokens for understanding and discrete tokens for generation, as shown in \cref{fig:teaser}(d).
This \textit{hybrid token} design enables easy transformation of a pretrained VLM into a UMM with a simple two-stage training:

\textit{(i) Unified Vision Tokenizer Training}:
\name trains a quantizer to discretize visual tokens and an image decoder to generate images.
The vision encoder is based on the \textit{frozen} pretrained VLM.
The continuous tokens from the VLM are for understanding and the discrete tokens from the quantizer are for generation.
With the encoder frozen, the vision tokenizer can be trained solely with image reconstruction loss, without requiring vision-language alignment~\cite{ma2025unitok}.

\textit{(ii) Unified Multimodal Training}:
\name optimizes the LLM for both image understanding and generation.
Instead of training from pretrained LLMs like prior UMMs \cite{wu2025janus,chen2025januspro,wu2024vilau}, we initialize the LLM from a pretrained VLM, enabling the model to start from the learned vision-language alignments rather than learning them from scratch.

We conduct a comprehensive evaluation on both multimodal understanding and image generation benchmarks, and compare \name with the latest UMMs, including Emu3~\cite{wang2024emu3}, Janus series~\cite{wu2025janus, chen2025januspro}, Show-o series~\cite{xie2024show_o, xie2025show_o2}, VILA-U~\cite{wu2024vilau}, and more.
\name achieves the best performance on multimodal understanding benchmarks such as GQA~\cite{hudson2019gqa}, POPE~\cite{li2023pope}, and VQAv2~\cite{antol2015vqa,goyal2017vqav2} and competitive performance on generation benchmarks, with a total cost of about \$2,000.\footnote{We estimate the cost of an 8$\times$H100 GPU node at \$20.3/hour and an 8$\times$A100 GPU node at \$8.56/hour according to \href{https://cloud-gpus.com/}{https://cloud-gpus.com/}.}
Our Argus-Unified-0.5B, one of the smallest UMMs using only 0.5B LLM, surpasses methods like Emu3~\cite{wang2024emu3} and Chameleon~\cite{team2024chameleon} with over 7B LLMs.
Notably, we use only 15.6M training data, significantly less than the existing works that use tens of millions to hundreds of millions, even billions of data (as shown in \cref{fig:performance}).

\section{Related Work}
\label{sec:related-work}

\subsection{Vision-language Models}

Vision-language models (VLMs) are multimodal models that process both visual and textual inputs~\cite{radford2021clip,liu2023llava,huang2026empirical}.
Pioneering works such as CLIP~\cite{radford2021clip} employ contrastive learning which takes natural language as supervision for visual representation learning \cite{zhuang2025argus}.
Recent VLMs, including LLaVA series~\cite{liu2023llava,liu2024llava,zhu2024llavaphi}, InternVL series~\cite{chen2024internvl2_5,zhu2025internvl3}, Qwen-VL series~\cite{bai2023qwenvl,bai2025qwen2_5vl}, commonly integrate a vision encoder with LLMs via a connector, \eg a multi-layer perceptron (MLP).
\cref{fig:teaser}(a) illustrates the high-level architecture of these VLMs.
The vision encoder---usually a Vision Transformer (ViT)~\cite{dosovitskiy2020vit} such as CLIP~\cite{radford2021clip}, SigLIP~\cite{zhai2023sigmoid}, or InternViT~\cite{chen2024internvl}---produces continuous visual tokens that are then aligned with textual embeddings.
Despite the advances of VLMs, they remain limited to text-based visual understanding and lack the ability of visual generation.

\subsection{Unified Multimodal Models}

Unified multimodal models (UMMs) aim to integrate both visual understanding and generation in a single model~\cite{lu2024unifiedio2,wu2024nextgpt,wang2024emu3,team2024chameleon}. Existing UMMs can be broadly categorized into three designs:

\textbf{Autoregressive-based UMMs} are generally built upon autoregressive (AR) LLMs and jointly model image and text generation.
Some works replace the vision encoder with that of a VQVAE~\cite{van2017vqvae} or VQGAN~\cite{esser2021vqgan} like tokenizer (\cref{fig:teaser}(b))~\cite{wang2024emu3,team2024chameleon,wu2024vilau,ma2025unitok,han2025tar,wu2024liquid}, while others incorporate an additional vision tokenizer dedicated for image generation (\cref{fig:teaser}(c))~\cite{wu2025janus,chen2025januspro,jiao2025unitoken,qu2025tokenflow}.

\textbf{Diffusion-based UMMs} employ diffusion models~\cite{rombach2022ldm,podell2023sdxl,sehwag2025stretching} instead of LLMs to generate images and texts~\cite{yang2025mmada,li2025ddit,shi2025muddit}.
However, they remain in early stages with suboptimal performance, partly due to the difficulty of mapping discrete text into a continuous space and capturing sequential dependencies.

\textbf{Hybrid AR+Diffusion UMMs} combine AR for text and diffusion for image generation.
Some works assemble a separate diffusion model on top of a pretrained VLM~\cite{tong2024metamorph,pan2025meta_queries}, forming loosely coupled architecture and separate token spaces
rather than a fully unified design.
Others integrate both paradigms within a transformer~\cite{ma2025janusflow,xie2024show_o,xie2025show_o2,wu2025harmonizing,fan2025unifluid}.
While diffusion offers high image fidelity, its independent training paradigm limits reuse of visual priors from pretrained AR-based VLMs.

Our \name is \textit{AR-based UMM} that effectively leverages pretrained VLMs and introduces a hybrid token design (\cref{fig:teaser}(d)), achieving the state-of-the-art performance in this paradigm.
Compared with other paradigms, it attains superior multimodal understanding performance and competitive generation quality with substantially lower training data and cost.

\section{Methodology}
\label{sec:method}

\begin{figure}[t!]
  \centering
  \includegraphics[width=1\linewidth]{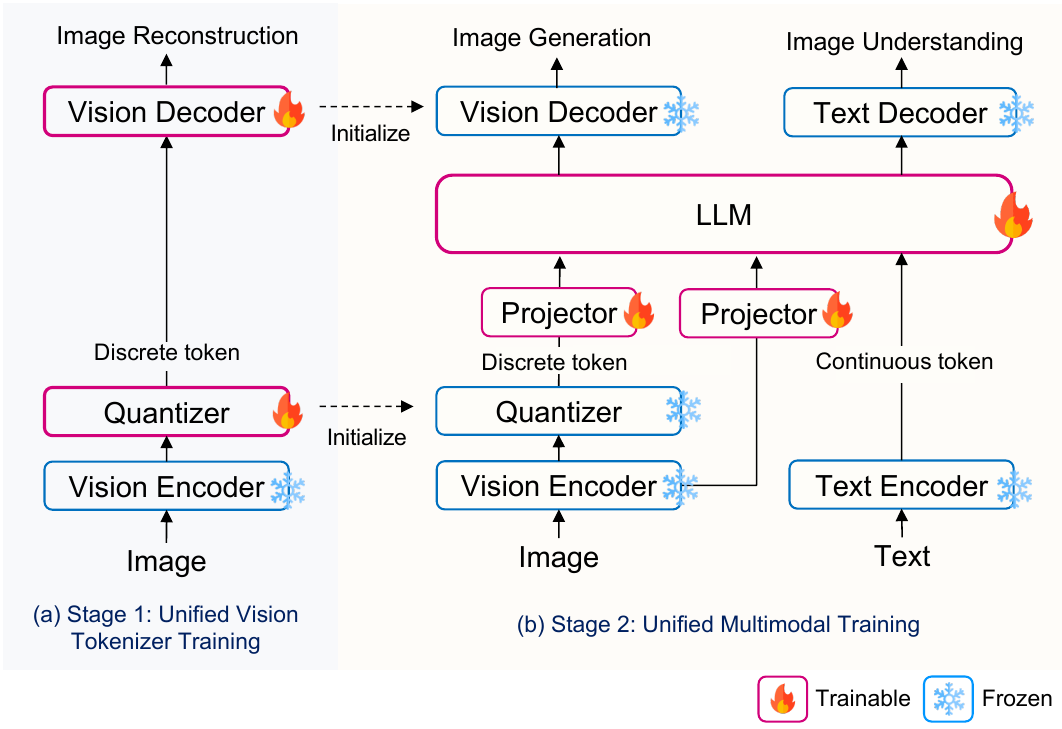}
  \caption{
Overview of \ours.
\ours transforms pretrained VLMs into UMMs via a novel hybrid token design and a two-stage training framework.
(a) In Stage 1, a quantizer and vision decoder are trained on top of the frozen vision encoder to enable it to produce both continuous tokens for understanding and discrete tokens for generation.
(b) In Stage 2, we initialize the model with pretrained VLM components and quantizer from Stage~1, then perform unified multimodal training to jointly optimize the LLM and projectors for understanding and generation.
  }\label{fig:overview}
\end{figure}

In this section, we present the design and training strategy of \name, a compact unified multimodal model (UMM) that performs both image understanding and image generation by effectively leveraging pretrained vision-language models (VLMs).
\name introduces a hybrid token design from a single vision encoder, which outputs both continuous tokens rich in semantics for understanding and quantized discrete tokens suitable for generation.
This design leverages the benefits of both token types within one encoder while avoiding parameter overhead.

\cref{fig:overview} provides an overview of the \name.
It consists of two major stages: (i) Unified Vision Tokenizer Training and (ii) Unified Multimodal Training.
In Stage~1 (\cref{fig:overview}(a)), \name builds a unified vision encoder by training a quantizer and decoder on top of the frozen vision encoder from a pretrained VLM.
It enables the vision encoder to produce both continuous tokens and discrete tokens for the next stage.
In Stage~2 (\cref{fig:overview}(b)), we initialize the model with pretrained VLM components and quantizer from Stage~1, then perform unified multimodal training to jointly optimize the LLM for both image understanding and generation.
Now, we describe these two training stages in detail.

\subsection{Unified Vision Tokenizer (Stage 1)}

We employ a shared vision encoder for image understanding and generation for architecture simplicity and model compactness.
Unlike prior UMMs that fine-tune the vision encoder~\cite{wu2024vilau,ma2025unitok,wu2025janus,chen2025januspro},
we \emph{freeze} a pretrained vision encoder from a strong VLM, preserving its vision-language alignments for understanding while efficiently extending it to image generation.
In comparison, prior UMMs with a unified vision encoder do not freeze their vision encoders~\cite{wang2024emu3,wu2024vilau,ma2025unitok},
as they need to simultaneously optimize discrete tokens for generation and align visual representations with texts.

The key to enabling autoregressive image generation lies in producing discrete visual tokens from the vision encoder.
We adopt the widely used VQVAE~\cite{van2017vqvae} and VQGAN~\cite{esser2021vqgan} architecture and trains a quantizer and an image decoder on top of the frozen pretrained vision encoder, as depicted in \cref{fig:overview}(a).
The quantizer discretizes the continuous embeddings from the vision encoder by mapping them to the nearest entries in a learned codebook of latent vectors, thus producing discrete codes that can be used as the prediction labels in autoregressive LLMs.
The decoder is trained to reconstruct the image from these discrete codes.

We train the quantizer and decoder with the following loss:
\begin{equation}
  \mathcal{L}
  = \mathcal{L}_{R}
  + \lambda_{P} \mathcal{L}_{P}
  + \lambda_{G} \mathcal{L}_{G}
  + \lambda_{\text{VQ}} \mathcal{L}_{\text{VQ}},
\end{equation}
where $\mathcal{L}_{R}$ is the pixel-wise reconstruction loss, $\mathcal{L}_{P}$ is the perceptual loss using LPIPS~\cite{zhang2018lpips}, $\mathcal{L}_{G}$ is the adversarial loss for enhancing reconstruction fidelity~\cite{karras2019style}, $\mathcal{L}_{\text{VQ}}$ is the vector quantization loss that minimizes distance between the vision encoder output and its nearest codebook entry~\cite{van2017vqvae}, and $\lambda$s are the weight coefficients for corresponding loss terms.
With the vision encoder frozen, our tokenizer is trained without the image-text loss~\cite{radford2021clip} used in prior works~\cite{wu2024vilau,ma2025unitok}.
This not only simplifies training---requiring only unlabeled image data---but also reduces computational costs.

\subsection{Unified Multimodal Training (Stage 2)}
\label{sec:stage2}

\cref{fig:overview}(b) illustrates the overall architecture of our \name. It consists of three main components: (1) the vision encoder, LLM, text encoder, and text decoder initialized from the pretrained VLM; 
(2) the quantizer and image decoder initialized from Stage~1; and
(3) two randomly initialized task-specific projectors for understanding and generation, respectively.
In this stage, the LLM and the two projectors are trainable.

\fakeparagraph{Hybrid Token Design.}
\name introduces hybrid tokens based on a unified vision encoder: continuous tokens for understanding and discrete visual tokens for generation.
Starting from a pretrained VLM vision encoder, we use its continuous visual representations for understanding, followed by a quantizer to produce discrete visual tokens for generation.
Both token types are projected into a shared LLM embedding space through separate projectors, enabling the model to handle modalities coherently.
Moreover, this design is cohesive with \emph{freezing} vision encoder in Stage~1, allowing \name to fully leverage pretrained VLMs  while extending them for image generation.
Prior UMMs mostly adopt discrete visual codes from VQGAN-like vision tokenizers~\cite{wu2024vilau,ma2025unitok} for both tasks, which limits the semantic richness for understanding.

\fakeparagraph{Training Objective.}
We train the LLM using the standard next-token prediction objective~\cite{radford2019gpt2,brown2020gpt3}.
Each sample is represented as a multimodal token sequence formed by concatenating visual tokens from the vision encoder and text tokens from the text tokenizer.
Specifically, the vision encoder converts an image into a 1D token sequence wrapped with special tokens \texttt{$<$image\_start$>$} and \texttt{$<$image\_end$>$}, inserted among text tokens following the format of chat template.
These multimodal tokens are fed into the LLM, producing text tokens for understanding and visual tokens for generation.
For image understanding, the model minimizes the negative log-likelihood (NLL) over text tokens.
For image generation, the output visual tokens are processed by a vision head and subsequently used to compute the NLL loss, similar to VILA-U~\cite{wu2024vilau}.
During inference, text tokens are decoded by the text decoder to generate text, while visual tokens are passed to the image decoder to generate image.

\footnotetext{For the works that do not report data scale, we tried our best to estimate it from the description of datasets used.}

\begin{table*}[t!]
\centering
\resizebox{\textwidth}{!}{%
\begin{tabular}{llllcccccc}
\toprule
\multirow{2}{*}{\textbf{Models}} & \multirow{2}{*}{\textbf{Type}}  & \multirow{2}{*}{\shortstack[l]{\textbf{LLM}\\\textbf{Scale}}} & \multirow{2}{*}{\shortstack[l]{\textbf{Data}\\\textbf{Scale\footnotemark}}} & \multicolumn{4}{c}{\textbf{Image Understanding}} & \multicolumn{2}{c}{\textbf{Image Generation}} \\
\cmidrule(l){5-10}
 &  &  &  & \textbf{GQA ↑} & \textbf{MME-P ↑} & \textbf{POPE ↑} & \textbf{VQAv2 ↑} & \textbf{MJHQ-30K ↓} & \textbf{GenEval ↑} \\
\midrule
MetaQuery~\cite{pan2025meta_queries} & AR+Diff. & 0.5B & 27.4M & - & 1238.0 & - & - & 6.3 & \textbf{0.74} \\
UniFork~\cite{li2025unifork} & AR & 0.76B & 82.3M & 55.1 & 1208.0 & 85.8 & 70.0 & 10.6 & 0.46 \\
Harmon~\cite{wu2025harmonizing} & AR+MAR & 0.5B & 109M & 56.3 & 1148.0 & 86.5 & - & \textbf{6.1} & 0.71 \\
\rowcolor{blue!5}\textbf{Argus-Unified (Ours)} & AR & 0.5B & \textbf{15.6M} & \textbf{61.4} & \textbf{1312.6} & \textbf{87.5} & \textbf{77.3} & 8.0 & 0.66 \\
\midrule
MMaDA~\cite{yang2025mmada} & Diffusion & 8B & 2B+ & 61.3 & 1410.7 & 86.1 & 76.7 & - & 0.63 \\
Seed-X~\cite{ge2024seedx} & AR+Diff. & 13B & 158M+ & 49.1 & 1457.0 & 84.1 & 71.2 & - & 0.51 \\
LaVIT~\cite{jin2023lavit} & AR+Diff. & 7B & 293M & 46.8 & - & - & 66.0 & - & - \\
ILLUME+ & AR+Diff. & 3B & 73M & - & 1414.0 & 87.6 & - & 6.0 & 0.53 \\
Chameleon~\cite{team2024chameleon} & AR & 34B & 1.4B & - & 604.5 & - & 69.6 & - & - \\
Chameleon~\cite{team2024chameleon} & AR & 7B & 1.4B & - & 202.7 & - & - & - & 0.39 \\
TokenFlow-L~\cite{qu2025tokenflow} & AR & 13B & 760M+ & 62.6 & 1365.4 & 85.0 & 73.9 & - & - \\
Emu3~\cite{wang2024emu3} & AR & 8B & - & 60.3 & 1243.8 & 85.2 & 75.1 & - & 0.66 \\
TokLIP-L~\cite{lin2025toklip} & AR & 7B & 125M & 59.5 & \textbf{1488.4} & 84.1 & - & - & - \\
LWM~\cite{liu2024lwm} & AR & 7B & 1B & 44.8 & - & 75.2 & 55.8 & 17.8 & 0.47 \\
SemHiTok-256~\cite{chen2025semhitok} & AR & 7B & 70M & 60.3 & 1449.0 & 83.4 & - & 5.4 & - \\
UniTok~\cite{ma2025unitok} & AR & 7B & 1B+ & 61.1 & 1448.0 & 83.2 & 76.8 & 7.5 & - \\
Liquid~\cite{wu2024liquid} & AR & 7B & 60M & 58.4 & 1119.3 & 81.1 & 71.3 & 5.5 & - \\
ViLA-U-256~\cite{wu2024vilau} & AR & 7B & 726M & 58.3 & 1336.2 & 83.9 & 75.3 & 12.8 & 0.41 \\
\addlinespace[2pt]
\cdashline{3-10}
\addlinespace[3pt]
D-Dit-512~\cite{li2025ddit} & Diffusion & 2B & 40M & 59.2 & 1124.7 & 84.0 & 60.1 & - & 0.50 \\
Show-o2~\cite{xie2025show_o2} & AR+Diff. & 1.5B & 66M & 60.0 & 1450.9 & - & - & - & 0.73 \\
Show-o-512~\cite{xie2024show_o} & AR+Diff. & 1.3B & 2B+ & 58.0 & 1097.2 & 80.0 & 69.4 & - & 0.68 \\
Harmon~\cite{wu2025harmonizing} & AR+MAR & 1.5B & 109M & 58.9 & 1155.0 & 87.6 & - & \textbf{5.2} & \textbf{0.76} \\
Janus-Pro~\cite{chen2025januspro} & AR & 1.5B & 234M & 59.3 & 1444.0 & 86.2 & - & - & 0.73 \\
Janus~\cite{wu2025janus} & AR & 1.3B & 100M+ & 59.1 & 1338.0 & 87.0 & 77.3 & 10.1 & 0.61 \\
\rowcolor{blue!5}\textbf{Argus-Unified (Ours)} & AR & 1.5B & \textbf{15.6M} & \textbf{62.8} & 1395.3 & \textbf{87.9} & \textbf{79.1} & 6.9 & 0.71 \\
\bottomrule
\end{tabular}
}
\caption{Performance comparison to the leading unified multimodal models (UMMs) on multimodal understanding and image generation benchmarks. Our model achieves the state-of-the-art performance among autoregressive (AR) models, excelling on both understanding and generation. Compared with diffusion-based and hybrid (AR+MAR or AR+Diffusion) UMMs, \name achieves the best multimodal understanding results on GQA, POPE, and VQAv2, while maintaining competitive generation performance with significantly less data.}
\label{tab:comparison}
\end{table*}


\fakeparagraph{Training Recipes.}
Our unified multimodal training consists of pretraining and supervised finetuning (SFT), designed to effectively adapt a pretrained VLM into a UMM.
Prior UMMs typically require large-scale data for multimodal pretraining to teach the LLM visual token modeling, \ie, vision-language alignment~\cite{wu2024vilau,wu2024liquid}.
Conversely, our LLM is initialized from a pretrained VLM and already exhibits strong image understanding and text generation abilities.
Therefore, we can warm up the LLM for visual token generation with substantially less data during pretraining.
Specifically, we use generation data JourneyDB (4.1M)~\cite{sun2023journeydb} and a smaller scale understanding data ShareGPT4V (1.25M)~\cite{chen2024sharegpt4v} to maintain modality alignment.
The total data amount in pretraining is significantly lower than prior works~\cite{wu2024liquid,ma2025unitok,wu2025harmonizing}, which use over 60M data.

In SFT, we further fine-tune the LLM on roughly balanced data of understanding and generation.
For image understanding, we use WiT~\cite{srinivasan2021wit}, ShareGPT4V~\cite{chen2024sharegpt4v}, VFLAN~\cite{chen2024allava-vflan}, ScienceQA~\cite{lu2022scienceqa}, and MGM-Instruct~\cite{li2024mgm}, totaling 5.47M samples.
For text-to-image generation, we use JourneyDB~\cite{sun2023journeydb}, BLIP3-o~\cite{chen2025blip3o}, and Echo-4o Instruct~\cite{ye2025echo}, totaling 4.23M image-text pairs.

Altogether, we curate 9.7M image-text pairs for unified multimodal training.
We find that in our setting of leveraging pretrained VLMs, expanding pretraining data beyond a moderate size offers diminishing or even negative returns, while enlarging SFT data can improve performance. More details are discussed in \cref{sec:ablations}.

\section{Experiments}
\label{sec:experiments}

In this section, we detail the experimental setup, compare \ours with existing UMMs, VLMs, and image generation models, and present comprehensive ablation studies.

\subsection{Experiment Setup}

\fakeparagraph{Datasets.}
We curated a total number of 15.6 million public data, consisting of 9.7M image-text pairs for Stage 2 (as discussed in \cref{sec:stage2}) and additional 5.9M images from CC3M~\cite{sharma2018conceptual}, DALL-E~3~\cite{egan2024dalle3data}, and DiffusionDB~\cite{wang2022diffusiondb}. Stage 1 training uses these images and the images from the image-text dataset, resulting in 14M images for vision tokenizer training. Note that we only utilize publicly available datasets without using any internal or proprietary data like prior works~\cite{wu2024vilau,ma2025unitok} to facilitate reproducibility.
As shown in \cref{fig:performance} and \cref{tab:comparison}, our total data volume is an order of magnitude smaller than that of existing UMMs, some of which require hundreds of millions to billions of samples.

\fakeparagraph{Evaluation Metrics.}
We evaluate the multimodal understanding capability of \name on standard benchmarks, including POPE~\cite{li2023pope}, VQAv2~\cite{antol2015vqa,goyal2017vqav2}, GQA~\cite{hudson2019gqa}, and MME~\cite{zhang2021mme}, .
We also assess the image generation capability on two complementary benchmarks: MJHQ-30K~\cite{li2024playground-mjhq} and GenEval~\cite{ghosh2023geneval}.
MJHQ-30K evaluates visual fidelity using the Fréchet Inception Distance (FID) computed over 30K images.
GenEval focuses on compositional reasoning and object consistency in generated images.

\fakeparagraph{Implementation Details.}
We focus on compact unified multimodal models and develop two variants: Argus-Unified-0.5B and Argus-Unified-1.5B.
Argus-Unified-0.5B is built on InternVL3-1B~\cite{zhu2025internvl3}, consisting of  Qwen2.5-0.5B~\cite{qwen2024qwen2_5} as the LLM and an InternViT-300M~\cite{chen2024internvl} as the vision encoder.
Argus-Unified-1.5B is built on InternVL3-2B~\cite{zhu2025internvl3}, consisting of Qwen2.5-1.5B~\cite{qwen2024qwen2_5} as the LLM and InternViT-300M~\cite{chen2024internvl} as the vision encoder. 
For generation, we resize images to 448$\times$448 to use the same vision encoder and adopt multi-codebook design for quantization inspired by UniTok~\cite{ma2025unitok}.
The image decoder is initialized from ViTamin-L/16~\cite{chen2024vitamin}.
Both understanding and generation projectors are single-layer MLPs.

\begin{table}[t!]
\centering
\resizebox{1\linewidth}{!}{%
\setlength{\tabcolsep}{8pt}
\begin{tabular}{lllll}
\toprule
\multirow{2}{*}{\textbf{Models}} & \textbf{LLM}   & \textbf{Vision Enc.} & \textbf{Data}  & \multirow{2}{*}{\textbf{GPU Hours}} \\
                                 & \textbf{Scale} & \textbf{Scale}       & \textbf{Scale} &                                     \\
\midrule
Janus-Pro~\cite{chen2025januspro} & 1.5B &  345M & 234M & 27648 $\times$  A100 \\
Janus~\cite{wu2025janus}                           & 1.3B           & 345M                 & 100M+           & 21504 $\times$ A100                 \\
VILA-U~\cite{wu2024vilau}                          & 7B             & 316M                 & 726M           & 20000 $\times$ A100                 \\
Harmon~\cite{wu2025harmonizing}                          & 1.5B           & 943M                 & 109M           & 6144 $\times$ A100                  \\
\rowcolor{blue!5}  \textbf{Argus-Unified}           & \textbf{1.5B}  & \textbf{304M}        & \textbf{15.6M}   & \textbf{800 $\times$ H100}         \\
\bottomrule
\end{tabular}
}
\label{tab:computation-comparison}
\caption{
Compute comparison with unified multimodal models.
\name is trained with 8 $\times$ NVIDIA H100 for 4.17 days (Stage 1 + Stage 2), requiring substantially less compute.
}
\end{table}

\subsection{Performance Comparison}

We compare \name{} with over 20 latest unified multimodal models (UMMs) with LLM scales from 0.5B to 34B in \cref{tab:comparison}.
Existing UMMs generally follow one of three design paradigms:
(1) autoregressive (AR) for both understanding and generation, e.g., Emu3\cite{wang2024emu3}, Janus~\cite{wu2025janus}, and VILA-U~\cite{wu2024vilau};
(2) diffusion-based for both understanding and generation (Diffusion), such as MMaDA~\cite{yang2025mmada} and D-Dit~\cite{li2025dual}; or
(3) hybrid of AR and Diffusion or MAR~\cite{li2024mar} (AR+Diff. or AR+MAR), which uses AR for understanding and diffusion or MAR modeling for generation—either by assembling an external diffusion model (e.g., MetaQuery~\cite{pan2025meta_queries}) or fusing diffusion or MAR modeling into AR (e.g., Show-o ~\cite{xie2024show_o,xie2025show_o2}).

\begin{figure*}[t!]
  \centering
  \includegraphics[width=\linewidth]{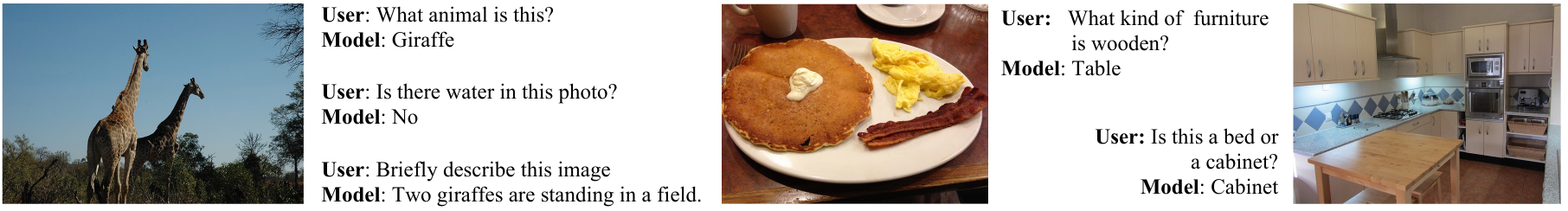}
  \caption{
  Image understanding examples produced by Argus-Unified-1.5B.
  }\label{fig:und-visualization}
\end{figure*}

\begin{figure*}[t!]
  \centering
  \includegraphics[width=\linewidth]{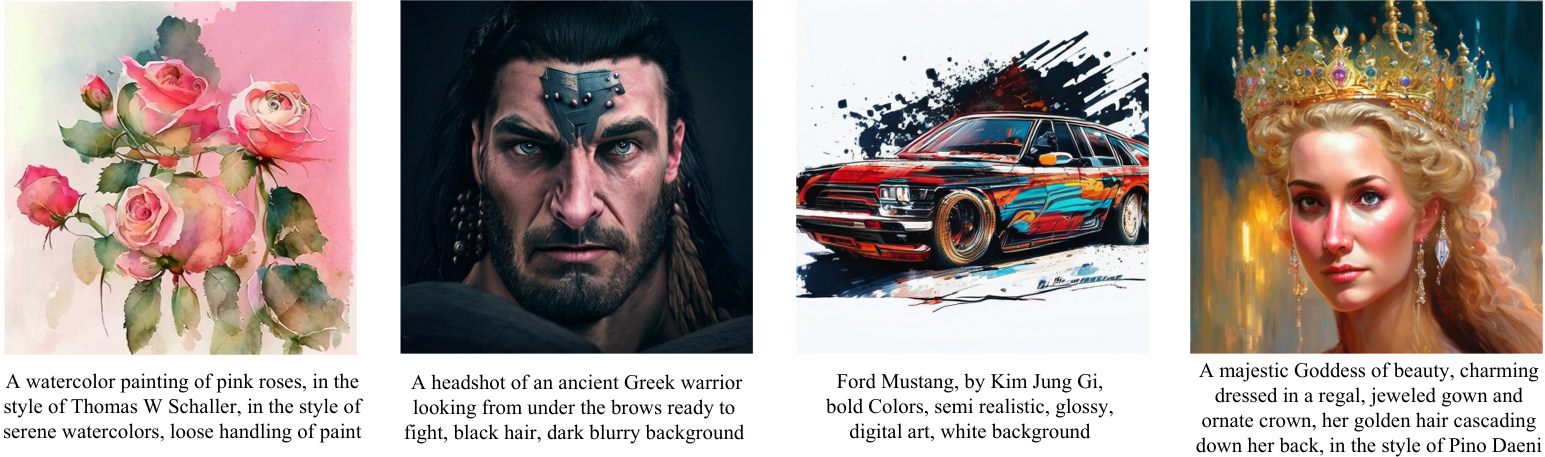}
  \caption{
  Images generated by Argus-Unified-1.5B. More examples are provided in the \cref{fig:gen-with-text}.
  }\label{fig:gen-visualization}
\end{figure*}

\fakeparagraph{Comparison with Unified Multimodal Models.} 
\name demonstrates strong performance across both understanding and generation benchmarks compared with existing UMMs.
Among AR UMMs,
\name achieves the state-of-the-art performance, surpassing representative works like Janus~\cite{wu2025janus} (which has a specialized encoder for generation) and VILA-U~\cite{wu2024vilau} across all benchmarks, and outperforming the other AR UMMs on majority of the benchmarks. Notably, Argus-Unified-1.5B even outperforms models with substantially larger LLMs (7B–13B) using far less data (only 15.6M).

Compared with diffusion-based and hybrid (AR+MAR or AR+Diffusion) UMMs, Argus-Unified-0.5B achieves the best results on all image understanding benchmarks among models with LLMs under 1B parameters and Argus-Unified-1.5B attains the best performance on GQA, POPE, and VQAv2, while remaining competitive on MME-P. For image generation, Argus-Unified-1.5B outperforms models with diffusion modeling such as Seed-X~\cite{ge2024seedx} and Show-o~\cite{xie2024show_o}, and achieves comparable results with Harmon~\cite{wu2025harmonizing} and MetaQuery~\cite{pan2025meta_queries}. Note that MetaQuery uses the same 0.5B-scale LLM but assembles an external diffusion model with approximately 1.6B parameters. Although Harmon-1.5B uses the same LLM scale as ours, its vision encoder with 943M parameters is about 3$\times$ larger than ours and it uses about 7$\times$ more data and more computation, as shown in \cref{tab:computation-comparison}. We also provide qualitative examples of understanding and generation in \cref{fig:und-visualization} and \cref{fig:gen-visualization}. More comparisons with models that only support image understanding or generation are provided in \cref{tab:comparison-und,tab:comparison-gen} in the Supplementary.

\begin{table}[t!]
\centering
\resizebox{0.48\textwidth}{!}{%
\setlength{\tabcolsep}{8pt}
\begin{tabular}{ccccccc}
\toprule
\multirow{2}{*}{\shortstack[m]{\shortstack[c]{\textbf{w/}\\\textbf{CFG}}}} & \multicolumn{4}{c}{\textbf{Image Understanding}}                       & \multicolumn{2}{c}{\textbf{Image Generation}} \\  \cmidrule(l){2-7}                                           & \textbf{GQA↑} & \textbf{MME-P↑} & \textbf{POPE↑} & \textbf{VQAv2↑} & \textbf{MJHQ↓}    & \textbf{GenEval↑}    \\
\midrule
$\times$                                    & \textbf{63.0}  & \textbf{1405.1}  & 87.7            & 79.1             & 8.2                   & \textbf{0.71}         \\
$\checkmark$                                & 62.8           & 1395.3           & \textbf{87.9}   & \textbf{79.1}    & \textbf{6.9}          & \textbf{0.71}        \\
\bottomrule
\end{tabular}
}
\caption{
Performance of Argus-Unified training with (w/) and without classifier-free guidance (CFG)~\cite{ho2022cfg}. 
CFG improves MJHQ-30K~\cite{li2024playground-mjhq} FID while preserving performance on the others.
}
\label{tab:cfg-impact}
\end{table}

\begin{table*}[t!]
\centering
\resizebox{0.8\textwidth}{!}{%
\setlength{\tabcolsep}{4pt}
\begin{tabular}{l|ccc|cc>{\columncolor{blue!5}}c>{\columncolor{blue!5}}c>{\columncolor{blue!5}}c}
\toprule
\textbf{Type}
& \multicolumn{3}{c|}{\textbf{Gen. Only}}
& \multicolumn{5}{c}{\textbf{Unified Models}} \\
\midrule

\textbf{Models}
& VQGAN
& RQ-VAE
& VAR
& TokenFlow
& VILA-U
& Argus-Unified
& \shortstack{Argus-Unified\\(Codebook$\times$2)}
& \shortstack{Argus-Unified\\(SigLIP2)} \\
\midrule

\textbf{Data Scale}
& -
& -
& -
& -
& 700M
& 14M
& 14M
& 14M \\

\textbf{ImageNet rFID ↓}
& 4.98
& 1.30
& 0.90
& 1.37
& 1.80
& 1.60
& 1.04
& 0.63 \\

\bottomrule
\end{tabular}
}
\caption{
Comparison of vision tokenizers on ImageNet~\cite{deng2009imagenet} reconstruction FID (rFID).
Argus-Unified achieves competitive performance with a frozen vision encoder and only 14M data. 
A larger codebook size ($\times$2) and alternative vision encoders (e.g., SigLIP2-Large \cite{tschannen2025siglip2}) can further improve results.
}
\label{tab:tokenizer}
\end{table*}

\begin{table*}[t]
\centering
\resizebox{0.8\textwidth}{!}{%
\setlength{\tabcolsep}{8pt}
\begin{tabular}{lcccccc}
\toprule
\multicolumn{1}{l}{\multirow{2}{*}{\textbf{Pretrained Weights}}} & \multicolumn{4}{c}{\textbf{Image Understanding}}                       & \multicolumn{2}{c}{\textbf{Image Generation}} \\\cmidrule(l){2-7}
\multicolumn{1}{c}{}                                             & \textbf{GQA ↑} & \textbf{MME-P ↑} & \textbf{POPE ↑} & \textbf{VQAv2 ↑} & \textbf{MJHQ-30K ↓}    & \textbf{GenEval ↑}    \\
\midrule
Pretrained ViT \& LLM                                            & 56.54          & 1361.51          & 85.03           & 75.26            & 8.77                  & 0.61                  \\
\textbf{Pretrained VLM}                                          & \textbf{62.95} & \textbf{1405.06} & \textbf{87.70}  & \textbf{79.10}   & \textbf{8.22}         & \textbf{0.71}        \\
\bottomrule
\end{tabular}
}
\caption{
Impact of initialization strategy on unified multimodal training.
Using pretrained VLM (InternVL3-2B) consistently outperforms initializing from separate pretrained ViT (InternViT) and LLM (Qwen2.5-1.5B).
}
\label{tab:pretrained-weights-comparison}
\end{table*}

\begin{table*}[t!]
\centering
\resizebox{0.78\textwidth}{!}{%
\setlength{\tabcolsep}{8pt}

\begin{tabular}{ccccccc}
\toprule
\multirow{2}{*}{\textbf{Token Type}}      & \multicolumn{4}{c}{\textbf{Image Understanding}}                       & \multicolumn{2}{c}{\textbf{Image Generation}} \\\cmidrule(l){2-7}
                                  & \textbf{GQA ↑} & \textbf{MME-P ↑} & \textbf{POPE ↑} & \textbf{VQAv2 ↑} & \textbf{MJHQ-30K ↓}    & \textbf{GenEval ↑}    \\
\midrule
Discrete Token              & 60.10          & 1257.43          & 85.73           & 73.23            & 8.34                  & \textbf{0.72}                  \\
Hybrid Token               & \textbf{62.95}          & \textbf{1405.06}          & \textbf{87.70}           & \textbf{79.10}            & \textbf{8.22}                  & 0.71                 \\
\bottomrule
\end{tabular}
}
\caption{
Comparison of discrete vs. hybrid token designs. Using continuous tokens for understanding and discrete tokens for generation (hybrid token) significantly enhances image understanding while preserving generation quality.
}
\label{tab:token-type}
\end{table*}

\subsection{Recipes and Ablation Studies}
\label{sec:ablations}

\fakeparagraph{Impact of Classifier-Free Guidance (CFG).}
CFG is widely used to improve generation quality and control how closely output follows a text prompt~\cite{ho2022cfg}.
We empirically study the impact of CFG in unified multimodal training and present the results in \cref{tab:cfg-impact}.
Using CFG in training effectively improves the FID score on MJHQ-30K~\cite{li2024playground-mjhq} while preserving the performance on GenEval~\cite{ghosh2023geneval} and understanding benchmarks.
These results demonstrate the effectiveness of CFG in the unified multimodal training.
For the rest of the ablation studies, we do not apply CFG to disentangle the impact of different components.


\fakeparagraph{Vision Tokenizer.}
We evaluate our vision tokenizer on ImageNet~\cite{deng2009imagenet} reconstruction FID (rFID) and compare it with recent tokenizers in \cref{tab:tokenizer}.
Despite the vision encoder is frozen and using only 14M images, ours achieves comparable ImageNet rFID to tokenizers designed for generation-only or unified models. We further visualize the reconstructed images in \cref{fig:reconstruct-visualization}. Notably, \ours with a frozen encoder significantly outperforms VILA-U~\cite{wu2024vilau}, which does not freeze the encoder.

In addition, we conduct ablations by doubling the codebook size and replacing the encoder with a pretrained SigLIP2-Large ~\cite{tschannen2025siglip2}. Both modifications lead to improved rFID and reconstruction quality. These results indicate that neither the tokenizer nor freezing the encoder freezing constitutes a bottleneck for unified models. Instead, our findings highlight the importance of effective training strategies in unified modeling, including classifier-free guidance (CFG), initialization from pretrained vision-language models (VLMs), and careful dataset design. We discuss these factors in greater detail in the subsequent ablation studies.

\fakeparagraph{Pretrained VLM vs. Pretrained ViT \& LLM.}
\cref{tab:pretrained-weights-comparison} compares initializing the model from a pretrained VLM (InternVL3-2B) versus separately pretrained vision encoder (InternViT) and language model (Qwen2.5-1.5B). Note that InternVL3-2B is built upon InternViT~\cite{chen2024internvl} and Qwen2.5-1.5B~\cite{qwen2024qwen2_5}, i.e., the architectures are identical across settings. We train both using the exact same pipeline and data. The results show that initializing from a pretrained VLM yields substantially stronger performance, demonstrating the benefit of leveraging pretrained VLMs.


\fakeparagraph{Token Type.} 
Most existing UMMs use discrete tokens for both understanding and generation. In contrast, we introduce a hybrid token design that uses continuous tokens for understanding and discrete tokens for generation. As shown in \cref{tab:token-type}, this hybrid design substantially enhances image understanding performance while preserving competitive results on image generation benchmarks. 
These results demonstrate the effectiveness of the hybrid token design in improving multimodal understanding without incurring additional costs.


\section{Conclusion}

In this work, we present \name, a compact and economical UMM that integrates visual understanding and generation.
\name leverages pretrained VLMs to inherit strong visual priors and employs a hybrid token design.
Extensive experiments demonstrate that \name achieves state-of-the-art performance on multiple understanding benchmarks and competitive generation quality, while requiring the least amount of data and computation cost. 
We hope that this work can serve as a practical baseline for future research on developing UMM at a low cost.

\section*{Limitations}
\name focuses on the compact regime, and its effectiveness at larger model scales remains an open question. In addition, while our models demonstrate strong performance on multimodal understanding and generation benchmarks, the evaluation does not explore other capabilities such as image editing, which we leave for future work.


\bibliography{custom}

\newpage
\clearpage
\appendix

\section{Implementation Details}

In this section, we provide more details on the model architecture, datasets, and experiment setup.

\subsection{Details on Model Architecture}

\fakeparagraph{Unified Vision Tokenizer.} Our tokenizer consists of a vision encoder and a quantizer. We propose a hybrid token design that unifies the vision encoder by producing continuous token for image understanding and discrete token for image generation.
The vision encoder is a vision transformer (ViT) \cite{dosovitskiy2020vit}, specifically InternViT \cite{chen2024internvl}. For Argus-Unified-1.5B, it is initialized from InternVL3-2B \cite{zhu2025internvl3}, and for Argus-Unified-0.5B, from InternVL3-1B \cite{zhu2025internvl3}. To remain consistent with the InternVL encoder, we split each image into $n$ tiles of size $S \times S$ (where $S = 448$). The model size is approximately 300M parameters. 
We use $n = 6$ tiles for understanding and no tiling for image generation in Stage~2 training~\cite{chen2024internvl2_5}. The encoder outputs continuous visual tokens, which follow two pathways: (1) they are directly fed into the LLM for image understanding tasks, and
(2) they are passed to the quantizer to obtain discrete tokens for image generation. For quantization, we adopt a multi-codebook design inspired by UniTok \cite{ma2025unitok}. We employ a lightweight transformer to infer multi-codebooks from LLM output visual tokens, as the final tokens input to the tokenizer decoder. Both continuous and discrete tokens are projected into the LLM’s feature space using separate MLP layers, one MLP layer for each token type.

\fakeparagraph{Unified LLM.} We use a standard decoder-only LLM to jointly train on mix image understanding and generation data.
For Argus-Unified-1.5B, the LLM is Qwen2.5-1.5B \cite{qwen2024qwen2_5}, initialized from InternVL3-2B \cite{zhu2025internvl3}.
For Argus-Unified-0.5B, the LLM is Qwen2.5-0.5B \cite{qwen2024qwen2_5}, initialized from InternVL3-1B \cite{zhu2025internvl3}.

\fakeparagraph{Image Decoder.} The image decoder is based on ViTamin-L/16~\cite{chen2024vitamin}, which combines transformer-based feature decoding with convolutional upsampling for spatial reconstruction. The decoder contains approximately 350M parameters.

\fakeparagraph{Training Pipeline.} As shown in \cref{fig:overview}, we train our models in two stages: unified vision tokenizer training and unified multimodal training. In Stage 1, we train the quantizer and the image decoder on top of the frozen vision encoder. In Stage 2, we train the LLM jointly with mixed understanding and generation data, while the image and text decoders are not involved in training.
The LLM output text tokens are processed by the LLM linear head before computing the negative log-likelihood (NLL) loss.
The LLM output visual tokens are further processed by a lightweight transformer head, similar to \cite{wu2024vilau,ma2025unitok}, before computing the NLL loss.
The final training objective is the sum of these two losses.

\fakeparagraph{Inference Pipeline.}
For image understanding, the input image is encoded by the vision encoder. The resulting continuous tokens are projected and concatenated with text tokens from the text encoder, and then fed into the LLM. The generated text tokens are decoded by the text decoder.
For image generation, the text prompt is fed into the LLM to autoregressively predict discrete visual tokens, which are then decoded by the image decoder to produce the final image.

\subsection{Details on Datasets}

\cref{tab:dataset-info} summarizes the datasets used to train \ours. Stage 1 uses images from all the datasets, totaling 14M images. Stage 2 pretraining (PT) uses only ShareGPT4V \cite{chen2024sharegpt4v} pretraining subset and JourneyDB \cite{sun2023journeydb}. Stage 2 supervised finetuning (SFT) uses 9.7M image-text pairs drawn from the following datasets: ShareGPT4V~\cite{chen2024sharegpt4v}, WiT~\cite{srinivasan2021wit}, VFLAN~\cite{chen2024allava-vflan}, ScienceQA~\cite{lu2022scienceqa}, and MGM-Instruct~\cite{li2024mgm}, JourneyDB~\cite{sun2023journeydb}, BLIP3-o~\cite{chen2025blip3o}, and Echo-4o Instruct~\cite{ye2025echo}.

\begin{table*}[h!]
\centering
\caption{
Summary of datasets used in \ours. $\checkmark$ indicates that the dataset is used in the corresponding training stage. Note that Stage 1 uses only the images from these datasets.
}
\label{tab:dataset-info}
\resizebox{0.9\textwidth}{!}{
\setlength{\tabcolsep}{8pt}
\begin{tabular}{lcccc}
\toprule
Datasets          & \multicolumn{1}{c}{Size (M)} & Stage 1 & Stage 2 PT & Stage 2 SFT \\
\midrule
ShareGPT4V PT \cite{chen2024sharegpt4v}    & 1.25                         & $\checkmark$       & $\checkmark$          &             \\
ShareGPT4V SFT \cite{chen2024sharegpt4v}  & 0.67                         & $\checkmark$       &            & $\checkmark$           \\
WiT \cite{srinivasan2021wit}             & 0.54                         & $\checkmark$       &            & $\checkmark$           \\
VFLAN \cite{chen2024allava-vflan}            & 1.50                         & $\checkmark$       &            & $\checkmark$           \\
ScienceQA \cite{lu2022scienceqa}        & 0.01                         & $\checkmark$       &            & $\checkmark$           \\
MGM-Instruct \cite{li2024mgm}     & 1.50                         & $\checkmark$       &            & $\checkmark$           \\
JourneyDB \cite{sun2023journeydb}              & 4.10                         & $\checkmark$       & $\checkmark$          & $\checkmark$           \\
BLIP3-o \cite{chen2025blip3o}          & 0.06                         & $\checkmark$       &            & $\checkmark$           \\
Echo-4o Instruct \cite{ye2025echo} & 0.07                         & $\checkmark$       &            & $\checkmark$           \\
CC3M \cite{sharma2018conceptual} & 2.92                         & $\checkmark$       &            &             \\
DALL-E 3 \cite{egan2024dalle3data}        & 1.01                         & $\checkmark$       &            &             \\
DiffusionDB \cite{wang2022diffusiondb}     & 2.00                         & $\checkmark$       &            &             \\
\midrule
Total            & 15.6                        &         &            &            \\
\bottomrule
\end{tabular}
}
\end{table*}

\begin{table*}[t!]
\centering
\caption{
Comparison with understanding-only models. 
Argus-Unified achieves superior performance on GQA~\cite{hudson2019gqa}, POPE~\cite{li2023pope}, and VQAv2~\cite{antol2015vqa,goyal2017vqav2}, while remaining competitive on MME-P~\cite{zhang2021mme}.
}
\label{tab:comparison-und}
\resizebox{0.8\textwidth}{!}{%
\setlength{\tabcolsep}{8pt}
\begin{tabular}{lccccc}
\toprule
\multirow{2}{*}{\textbf{Model}} & \textbf{LLM}      & \multirow{2}{*}{\textbf{GQA↑}} & \multirow{2}{*}{\textbf{MME-P↑}} & \multirow{2}{*}{\textbf{POPE↑}} & \multirow{2}{*}{\textbf{VQAv2↑}} \\
                                & \textbf{Scale} &                                &                                  &                                 &                                  \\
\midrule
MobileVLM~\cite{chu2023mobilevlm}                       & 1.4B              & 56.1                           & 1196.2                           & 84.5                            & -                                \\
MobileVLM-V2~\cite{chu2024mobilevlmv2}                   & 1.4B              & 59.3                           & 1302.8                           & 84.3                            & -                                \\
MiniGemini~\cite{li2024minigemini}             & 2B                & 59.9                           & 1341.0                           & 83.9                            & -                                \\
MobileVLM~\cite{chu2023mobilevlm}                      & 2.7B              & 59.0                           & 1288.9                           & 84.9                            & -                                \\
MobileVLM-V2~\cite{chu2024mobilevlmv2}                  & 2.7B              & 61.1                           & 1440.5                           & 84.7                            & -                                \\
LLaVA-Phi~\cite{zhu2024llavaphi}                     & 2.7B              & -                              & 1335.1                           & 85.0                            & 71.4                             \\
LLaVA~\cite{liu2023llava}                         & 7B                & -                              & 809.6                            & 76.3                            & -                                \\
LLaVA-v1.5~\cite{liu2024llava}                    & 7B                & 62.0                           & \textbf{1510.7}                  & 85.9                            & 78.5                             \\
InstructBLIP~\cite{dai2023instructblip}                  & 7B                & 49.2                           & -                                & -                               & -                                \\
Qwen-VL-Chat~\cite{bai2023qwenvl}                    & 7B                & 57.5                           & 1487.5                           & -                               & 78.2                             \\
IDEFICS-9B~\cite{laurenccon2023idefics}                     & 8B                & 38.4                           & -                                & -                               & 50.9                             \\
InstructBLIP~\cite{dai2023instructblip}                  & 13B               & 49.5                           & 1212.8                           & 78.9                            & -                                \\
\rowcolor{blue!5} \textbf{Argus-Unified}          & 0.5B              & 61.2                           & 1323.2                           & 87.7                            & 77.4                             \\
\rowcolor{blue!5} \textbf{Argus-Unified}          & 1.5B              & \textbf{63.0}                  & 1405.1                           & \textbf{87.7}                   & \textbf{79.1}                    \\
\bottomrule
\end{tabular}
}

\end{table*}

Besides the main training datasets, we also use additional datasets for ablation studies. To examine the impact of data volume in Stage 2 pretraining, we increase the data amount to 16.17M by incorporating additional datasets: a subset of MMC4 Core (3.8M) \cite{zhu2023mmc4}, CC3M (2.9M) \cite{sharma2018conceptual}, a subset of CC12M (2.9M) \cite{changpinyo2021cc12m}, a subset of DALL-E 3 data (0.46M) \cite{egan2024dalle3data}, and a subset of DiffusionDB (0.76M) \cite{wang2022diffusiondb}. These additional datasets sum to 10.82M. Combined with ShareGPT4V PT (1.25M) \cite{chen2024sharegpt4v} and JourneyDB (4.1M) \cite{sun2023journeydb}, the total pretraining data volume is 16.17M.

In addition, we conduct ablation on the Stage 1 training data by using 51.2M of DataComp \cite{gadre2023datacomp}. More details are provided in \cref{sec:more-experiments}.

\begin{table*}[]
\centering
\caption{
Comparison with generation-only models. Argus-Unified achieves superior performance on MJHQ-30K~\cite{li2024playground-mjhq} and GenEval~\cite{ghosh2023geneval} benchmarks.
}
\label{tab:comparison-gen}
\resizebox{0.8\textwidth}{!}{%
\setlength{\tabcolsep}{8pt}
\begin{tabular}{llccc}
\toprule
\multirow{2}{*}{\textbf{Models}} & \multirow{2}{*}{\textbf{Type}}       & \multirow{2}{*}{\textbf{\begin{tabular}[c]{@{}c@{}}Model\\Scale\end{tabular}}} & \textbf{MJHQ-30K}     & \textbf{GenEval}  \\
                                 &           &                                                                                     & \textbf{FID↓}         & \textbf{Overall↑} \\
\midrule
LlamaGen~\cite{sun2024llamagen}                         &  AR         & 0.8B                                                                                & 25.6                 & 0.32              \\
LDM~\cite{rombach2022ldm}                              &    Diffusion       & 0.4B                                                                & - & 0.37              \\
SD v1.5 ~\cite{rombach2022ldm}                        & Diffusion & 1B                                                                                  & - & 0.43              \\
PixArt-$\alpha$~\cite{chen2023pixart}                        & Diffusion & 0.6B                                                                                & -                  & 0.48              \\
SD v2.1~\cite{rombach2022ldm}                         & Diffusion & 1B                                                                                  & 27.0                 & 0.50              \\
SD-XL~\cite{podell2023sdxl}                            & Diffusion & 2.6B                                                                                & 8.8                  & 0.55              \\
DALL-E 3~\cite{betker2023dalle3}                        & Diffusion & -                                                                & - & 0.67              \\
\rowcolor{blue!5} \textbf{Argus-Unified}           & AR        & 0.5B                                                                                & 8.0                   & 0.66              \\
\rowcolor{blue!5} \textbf{Argus-Unified}           & AR        & 1.5B                                                                                & \textbf{6.9}                   & \textbf{0.71}             \\
\bottomrule
\end{tabular}
}
\end{table*}

\begin{table*}[t!]
\caption{
Comparison of data volume and type in unified multimodal training with a pretrained VLM. More supervised finetuning (SFT) data improves performance (row 1 vs 2), while excessive pretraining (PT) data (row 3) or generation-only PT data (row 4) degrades performance. These results highlight the insights of constructing data in unified multimodal training with pretrained VLMs.
}
\label{tab:data-volume-type}
\centering
\resizebox{1\textwidth}{!}{%
\setlength{\tabcolsep}{8pt}
\begin{tabular}{lllcccccc}
\toprule
\multicolumn{2}{c}{\textbf{Datasets (Volume)}} & & \multicolumn{4}{c}{\textbf{Image Understanding}}                       & \multicolumn{2}{c}{\textbf{Image Generation}} \\\cmidrule(l){1-2}\cmidrule(l){4-9}
\multicolumn{1}{c}{\textbf{PT}}            & \multicolumn{1}{c}{\textbf{SFT}}          & & \textbf{GQA ↑} & \textbf{MME-P ↑} & \textbf{POPE ↑} & \textbf{VQAv2 ↑} & \textbf{MJHQ-30K ↓}    & \textbf{GenEval ↑}    \\
\midrule
und+gen (5.35M)        & und+gen (4.77M)       & & 62.0           & 1283.1           & 87.8            & 77.6             & 8.7                   & 0.30                  \\
und+gen (5.35M)        & und+gen (9.58M)       & & \textbf{63.0}           & \textbf{1405.1}           & 87.7            & \textbf{79.1}             & \textbf{8.2}                   & \textbf{0.71}                  \\
und+gen (16.17M)       & und+gen (9.58M)       & & 62.5           & 1314.9           & \textbf{88.1}            & 78.6             & 8.0                   & 0.68                  \\
\midrule
gen (4.1M)             & und+gen (9.58M)       & & 62.6           & 1379.1           & 87.1            & 78.8             & 8.3                   & 0.68                 \\
\bottomrule
\end{tabular}
}
\end{table*}

\subsection{Details on Experiment Setup}

All the experiments are conducted in one node of 8 NVIDIA H100 GPUs. By using the pretrained InternViT from InternVL3 \cite{zhu2025internvl3}, we follow similar image preprocessing to be consistent with the encoder design for image understanding. In particular, an input image is first resized it to the closest aspect ratio from a predefined set. The image is then partitioned into $n$ square tiles, each of which is further resized match the input resolution $S \times S$ of the vision transformer. The original input image is also resized to the same resolution, regarded as a thumbnail. Each tile and the thumbnail are independently encoded by the ViT into sequences of visual patch tokens. To reduce the total number of tokens, features from spatially adjacent patches are concatenated along the channel dimension. In our experiments, we use the same resolution $S = 448$ as InternVL3 \cite{zhu2025internvl3} vision encoder. We set $n = 6$ tiles for image understanding and no tiling for image generation.

By default, Stage 1 training is conducted for two epochs on 14M images with a batch size of 128. Stage 2 uses a batch size of 32 and a learning rate of $5 \times 10^{-5}$ during pretraining, and a batch size of 64 with a learning rate of $1 \times 10^{-4}$ during supervised finetuning. Both pretraining and supervised finetuning are run with 1 epoch.

\section{More Experiment Results}
\label{sec:more-experiments}

\fakeparagraph{Comparison with Existing Understanding-only Models.}
We further compare \name with vision–language models designed solely for image understanding, including the MobileVLM series~\cite{chu2023mobilevlm,chu2024mobilevlmv2} and the LLaVA series~\cite{liu2023llava,liu2024llava,zhu2024llavaphi}, as shown in \cref{tab:comparison-und}. Argus-Unified-1.5B achieves the best performance on GQA, POPE, and VQAv2 benchmarks among models with LLM scales ranging from 1.4B to 13B. Moreover, compared with MobileVLM and MobileVLM-V2 (both using a 1.4B LLM), even our smallest variant (0.5B LLM) delivers substantially stronger performance. These results highlight the strong image-understanding ability of Argus-Unified.

\begin{figure}[t!]
  \centering
  \begin{subfigure}[t]{0.45\textwidth}
      \includegraphics[width=\textwidth]{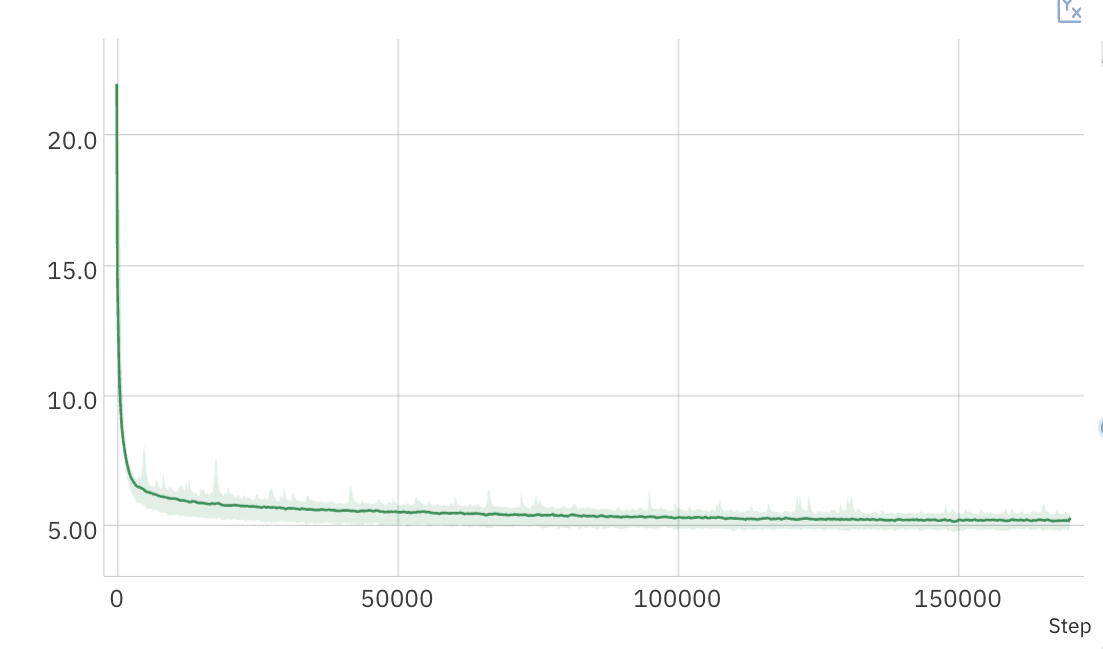}
      \caption{Pretraining}
      \label{fig:loss-pt}
  \end{subfigure}
  \begin{subfigure}[t]{0.45\textwidth}
    \includegraphics[width=\textwidth]{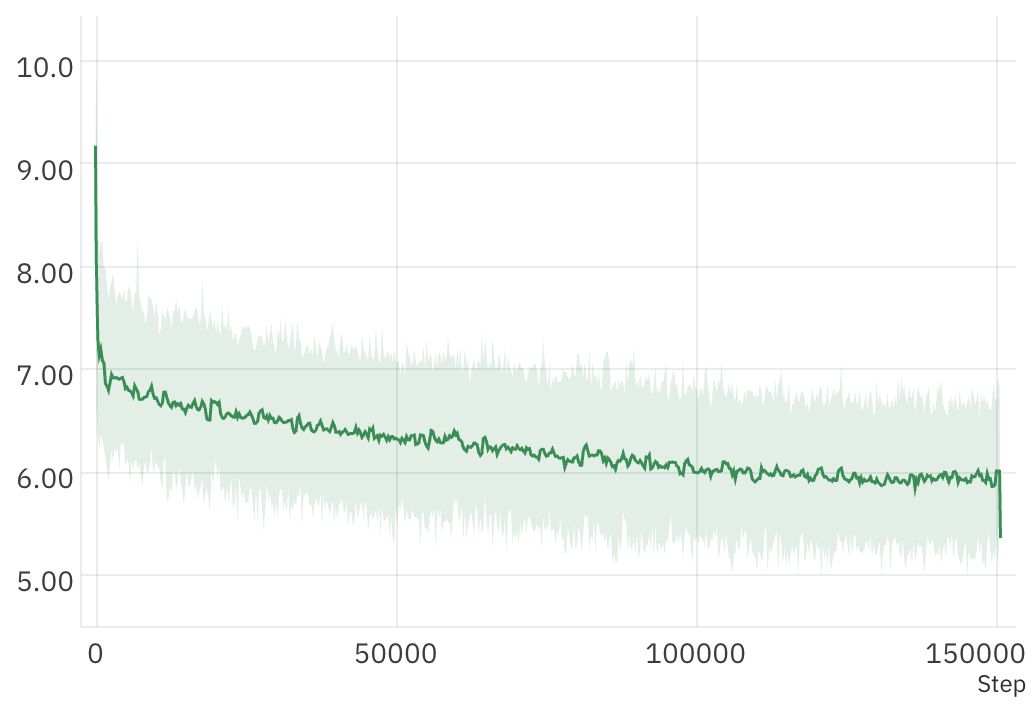}
    \caption{Supervised finetuning}
      \label{fig:loss-sft}
  \end{subfigure}
  \hfill
 \caption{Loss curves over the course of Stage 2 training.}
 \label{fig:loss}
\end{figure}

\begin{figure*}[t!]
  \centering
  \includegraphics[width=\linewidth]{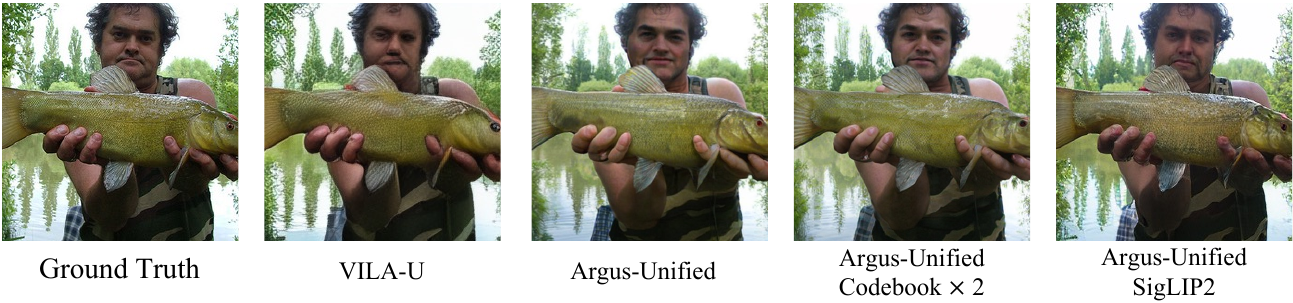}
  \caption{
  Comparison of reconstructed images on ImageNet. Our frozen-encoder design produces high-fidelity reconstructions and seamlessly generalizes to larger codebook and alternative vision encoders such as SigLIP2-Large \cite{tschannen2025siglip2}.
  }\label{fig:reconstruct-visualization}
\end{figure*}

\begin{table*}[t!]
\centering
\caption{
Ablation of Stage-1 tokenizer training data. Training the tokenizer on 51.2M of DataComp \cite{gadre2023datacomp} images improves ImageNet reconstruction (i.e., lower rFID) but yields weaker unified model performance than our mixed 14M data, indicating that reconstruction quality does not directly predict generation or understanding performance. It also highlights the effectiveness of our dataset curation for Stage 1 training.
}
\label{tab:tokenizer-argus}
\resizebox{\textwidth}{!}{%
\setlength{\tabcolsep}{4pt}
\begin{tabular}{lcccccccc}
\toprule
\multirow{2}{*}{\textbf{Stage 1 Data}} & \textbf{Image Reconstruction} & & \multicolumn{4}{c}{\textbf{Image Understanding}}                       & \multicolumn{2}{c}{\textbf{Image Generation}} \\\cmidrule{2-2}\cmidrule{4-9}
                                       & \textbf{ImageNet ↓} & & \textbf{GQA ↑} & \textbf{MME-P ↑} & \textbf{POPE ↑} & \textbf{VQAv2 ↑} & \textbf{MJHQ-30K ↓}    & \textbf{GenEval ↑}    \\
\midrule
Mixed-14M                                    & 1.60 & & \textbf{62.95} & \textbf{1405.06}          & \textbf{87.70}           & 79.10   & \textbf{8.22}         & \textbf{0.71}                  \\
DataComp-51.2M                               & 1.15  & & 62.37          & 1384.09 & 87.47  & \textbf{79.29}            & 8.68                  & 0.69        \\
\bottomrule
\end{tabular}
}
\end{table*}

\begin{table*}[t!]
\centering
\caption{
Comparison of image generation output resolutions. Higher resolution ($448\times448$) improves both image understanding and generation performance compared to $256\times256$.
}
\label{tab:output-resolution}
\resizebox{0.8\textwidth}{!}{
\setlength{\tabcolsep}{4pt}
\begin{tabular}{cccccccc}
\toprule
\multirow{2}{*}{\textbf{Output Resolution}} & \multicolumn{4}{c}{\textbf{Image Understanding}}                       & \multicolumn{2}{c}{\textbf{Image Generation}}          \\\cmidrule(l){2-7}
                                            & \textbf{GQA ↑} & \textbf{MME-P ↑} & \textbf{POPE ↑} & \textbf{VQAv2 ↑} & \textbf{MJHQ-30K ↓}    & \textbf{GenEval ↑}     \\
\midrule
256 $\times$ 256                                     & 62.65          & 1365.53          & 87.22           & 79.25            & 8.58                  & 0.71                  \\
448 $\times$ 448                                     & 62.95          & 1405.06          & 87.70           & 79.10            & 8.22                  & 0.71                           \\
\bottomrule
\end{tabular}
}
\end{table*}

\fakeparagraph{Comparison with Existing Generation-only Models.}
We also compare \name with representative image generation models, including Stable Diffusion models~\cite{rombach2022ldm,podell2023sdxl}, DALL-E~3~\cite{betker2023dalle3}, and the autoregressive model LlamaGen~\cite{sun2024llamagen}. As shown in \cref{tab:comparison-gen}, Argus-Unified-1.5B achieves an overall accuracy of 71\% on GenEval~\cite{ghosh2023geneval}  and an FID of 6.9 on MJHQ-30K~\cite{li2024playground-mjhq}, outperforming these popular generation models on both benchmarks. Our smaller variant (0.5B) also delivers notably strong results, demonstrating the high quality of images produced by Argus-Unified and the effectiveness of our training pipeline.

\fakeparagraph{Data Volume and Data Type.}  
We further investigate how to better utilize data when leveraging a pretrained VLM for UMM. Prior works typically adopt larger data volume for pretraining (PT) than for supervised fine-tuning (SFT)~\cite{wu2024liquid,ma2025unitok}. However, this trend does not hold when initializing from a pretrained VLM. As shown in \cref{tab:data-volume-type}, increasing the amount of SFT data improves performance (row 1 vs. row 2), but enlarging the PT data size from roughly 5M to 16M degrades performance on most benchmarks (row 2 vs. row 3). This suggests that the pretrained VLM already provides strong visual priors, and PT mainly serves as a warm-up phase for adapting to visual token generation. Moreover, we further study whether understanding data is needed by using only generation data in PT. This leads to consistently worse performance across all benchmarks (row 2 vs. row 4). These findings highlight the insights in constructing data when leveraging pretrained VLMs.

\fakeparagraph{Loss Curve Over the Course of Training.}
\cref{fig:loss} shows the loss curves for Stage 2 pretraining and supervised finetuning over the course of training. In both stages, the loss continues to decrease steadily throughout training.
It demonstrates that our training is stable and well-behaved across both pretraining and finetuning.

\fakeparagraph{Vision Tokenizer.}
We further conduct ablation study on unified vision tokenizer training by training another tokenizer using 51.2M images from DataComp-1B~\cite{gadre2023datacomp} (where UniTok~\cite{ma2025unitok} trains on the full dataset).  As shown in \cref{tab:tokenizer-argus}, although using 51.2M data achieves a lower ImageNet reconstruction rFID, using it for unified training leads to comparable but slightly worse performance compared to our default tokenizer trained on mixed 14M set. All Stage-2 training settings are exactly the same. This indicates shows that better reconstruction quality does not necessarily translate into stronger generation ability, and highlights the effectiveness of our dataset construction for Stage 1 unified vision tokenizer training.

\fakeparagraph{Generation Output Resolution.} Our framework can flexibly support different image generation output resolutions. We compare the performance of using resolution $256 \times 256$ and $448 \times 448$ in \cref{tab:output-resolution}. With strong performance obtained with resolution $256 \times 256$, using higher resolution can further increase the performance on both image understanding and image generation benchmarks. By default, we use the output resolution of $448 \times 448$ in our experiments.

\begin{figure*}[t!]
  \centering
  \begin{subfigure}[t]{\textwidth}
      \includegraphics[width=\textwidth]{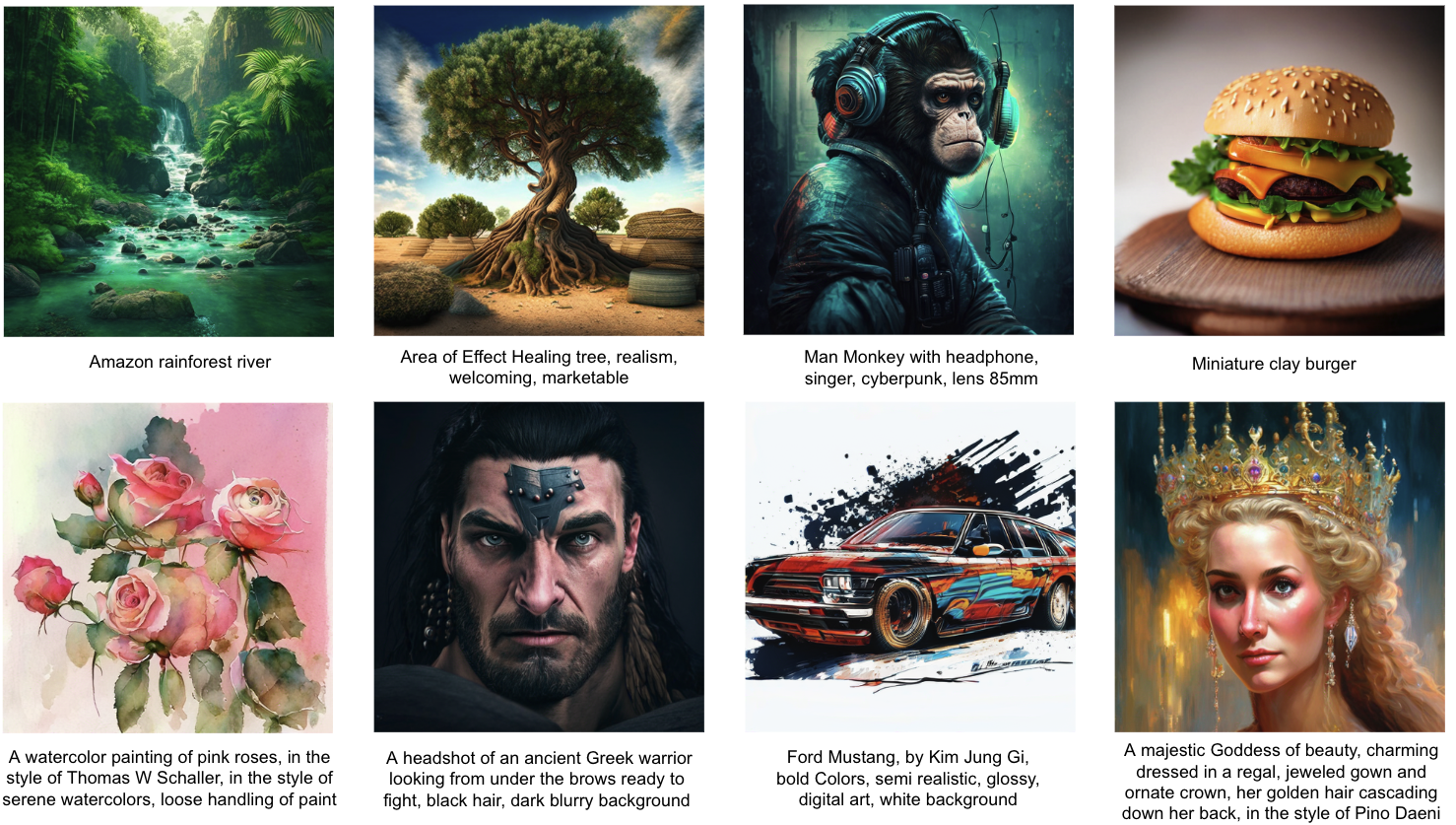}
  \end{subfigure}
  \begin{subfigure}[t]{\textwidth}
    \includegraphics[width=\textwidth]{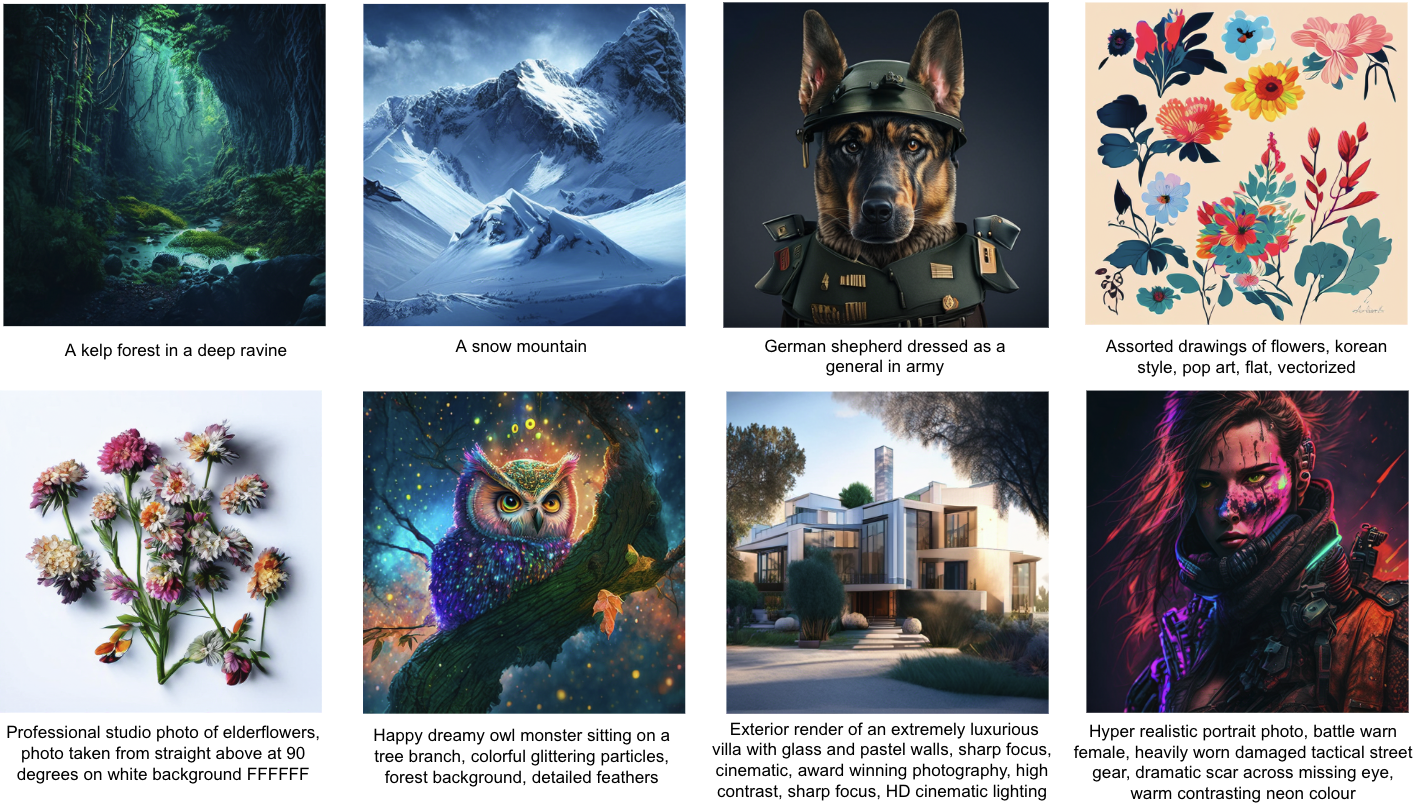}
  \end{subfigure}
  \hfill
 \caption{Images generated by \name{} from the corresponding text prompts.}
 \label{fig:gen-with-text}
\end{figure*}

\fakeparagraph{More Image Generation Results.}
To complement the visual examples in \cref{fig:gen-visualization}, we additionally include their corresponding prompts and more generated samples in \cref{fig:gen-with-text}.

\end{document}

%% file: preamble.tex
\def\ours{Argus-Unified\xspace}

\newcommand{\rank}[2][1]{\ifthenelse{\equal{#1}{1}}{\textbf{#2}}{\ifthenelse{\equal{#1}{2}}{\underline{#2}}{#2}}}
